%% file: final.tex
\input{_constants}
\documentclass[10pt,twocolumn,letterpaper]{article}
\usepackage{iccv}
\usepackage{epsfig}
\usepackage{graphicx}
\usepackage{amsmath}
\usepackage{amssymb}
\usepackage{booktabs}
\usepackage{multirow}

\usepackage{subcaption}
\usepackage[font={small}]{caption}
\usepackage[numbers, sort]{natbib}
\usepackage{hyperref}

\usepackage{cleveref}
\iccvfinalcopy % *** Uncomment this line for the final submission

\def\iccvPaperID{6064} % *** Enter the ICCV Paper ID here

\ificcvfinal\fi

\begin{document}

%%%%%%%%% TITLE
\title{\paperTitle}

\author{  
%%%\vspace{0.4em}
   Muzhi Zhu$^{1}$,~~
   Hengtao Li$^{1}$, ~~
   Hao Chen$^{1}$\thanks{HC is the corresponding author. WM was visiting Zhejiang University.}, ~~~
   Chengxiang Fan$^{1}$, ~ ~
   Weian Mao$^{2,1}$, ~~
   Chenchen	Jing$^{1}$, ~ \\ 
   {Yifan	Liu}$^{2}$, ~~
   {Chunhua Shen}$^{1}$ 
   \\[.2cm]
\normalsize   $^{1}$ Zhejiang University, China ~~~~~~
\normalsize  $^{2}$ The University of Adelaide, Australia 
}

\maketitle
\ificcvfinal\thispagestyle{empty}\fi

%%%%%%%%% ABSTRACT
\begin{abstract}

Current closed-set instance segmentation models rely on pre-defined class labels for each mask during training and evaluation,
largely 
limiting their ability to detect novel objects. Open-world instance segmentation (OWIS) models address this challenge by detecting unknown objects in a class-agnostic manner. However, previous OWIS approaches completely erase category information during training
to keep the model's ability to generalize to unknown objects. In this work, we propose a novel training mechanism %called
termed 
{\bf \handle} that %utilizes
uses 
category information to improve the model's class-agnostic segmentation ability for both known and unknown categories. In addition, 
the previous OWIS training setting exposes the unknown classes to the training set  and brings information leakage, which is unreasonable in the real world. 
Therefore, we provide a new open-world benchmark closer to a real-world scenario by dividing the dataset classes into known-seen-unseen parts. 
For the first time, we focus on the model's ability to discover objects that never appear in the training set images.

Experiments show that \handle can improve the overall and unseen detection performance by 5.6\% and 6.1\% in AR on our new benchmark without affecting the inference efficiency. We further demonstrate the effectiveness of our method on existing cross-dataset transfer and strongly supervised settings, leading to 5.5\% and 12.3\% relative improvement. 
Code and data are released at:

\url{https://github.com/aim-uofa/SegPrompt}

\end{abstract}

%%%%%%%%% BODY TEXT
\section{Introduction}
\label{sec:intro}

Datasets %has been
are 
one of the most important driving force for deep learning. For general foundation models, massive paired text-image data harvested from the internet is a convenient and valuable resource. However, for instance-level perception tasks such as detection and segmentation, it is
still 
very challenging to collect datasets at similar scales because every instance requires pixel-level %and category
annotations. Increasing the number of categories will introduce ambiguity and cause unrealistic manual labor. Category-level annotations prevail in these tasks, and an object can be discovered only if it is classified into one of the known classes in the training set. One promising solution is to decouple detection and classification tasks to increase the semantic complexity. For example, in open-vocabulary segmentation \cite{huynh2022open,ding2022open,xu2021simple}, a class-agnostic segmentation network is often adopted to locate objects, and the classification part relies on a vision language model. Under this framework, the ability to discover novel objects becomes crucial. In other words, we need to figure out how %we can
to 
discover more novel objects given the appropriate annotations. This task fits into the definition of open-world instance segmentation (OWIS) \cite{wang2021unidentified}, where the model is required to not only segment the  known categories but also the unknown categories. 

Recent work \cite{qi2021open} has shown that class-agnostic training encourages cross-dataset generalization. Previous work on OWIS also has agreed that class-aware training harms the model's ability to generalize to unknown objects.
%so
Thus, 
they all completely 
%drop
discard 
category information in the training phase. Consistent with this, our preliminary experimental results indicate that a stronger closed-world model generalizes better to unseen classes and class-agnostic training further improves the results. Specifically, Mask2Former \cite{cheng2022masked} surpasses Mask R-CNN by 3.9\% AR and class-agnostic training leads to another 2.5\% AR improvement.  All phenomena show that the direct inclusion of category information damages the generalization ability of the model.
\begin{figure}[tp]
    \centering
    \resizebox{1\columnwidth}{!}{
    \includegraphics[width=0.9\linewidth]{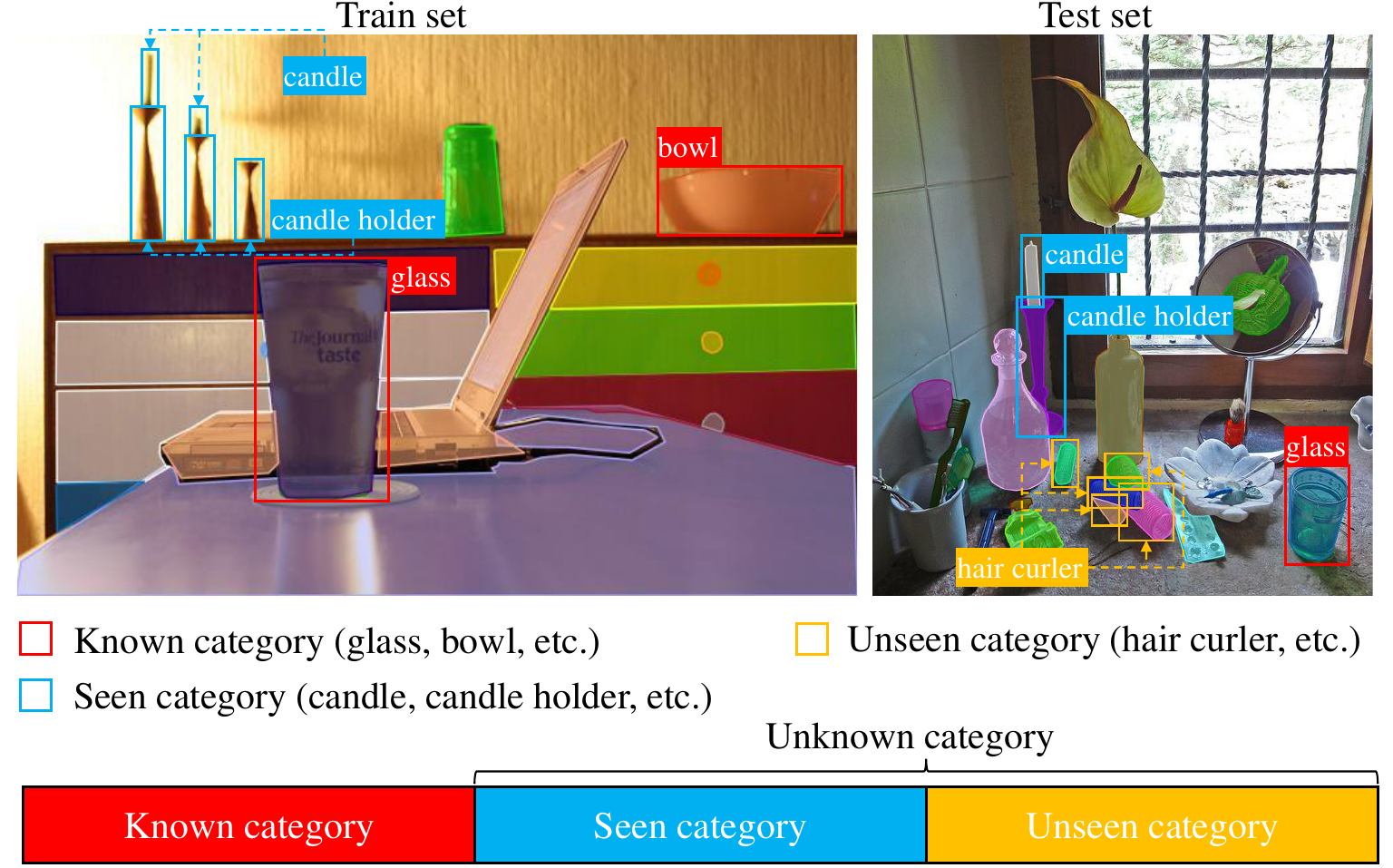}
    }
    \caption{\textbf{Our new benchmark, LVIS-OW.} 
    We divide the object categories into three parts and mark them with different colored boxes. `Known' represents categories that are labeled in the training set, `Seen' represents categories that appear in the training set but are not labeled, and `Unseen' represents categories that never appear in the training set. 
    }
    %%\vspace{-2em}
    \label{fig:setting}
\end{figure}

 In this work, we explore an interesting question: \textit{is the category information really useless}? 
 To make use of the category information while maintaining the generalization ability, we propose a prompt learning mechanism---{\bf \handle}---which serves as a training time auxiliary supervision to improve class-agnostic segmentation quality in general.
 The idea is to enhance category-level appearance representation learning while keeping the class-agnostic branch unaffected. 
The loose analogy can be drawn between our method and visual prompt tuning (VPT) \cite{jia2022visual}. While VPT is an efficient tuning protocol to transfer pre-trained models to downstream tasks, \handle can be viewed  as an efficient representation learning strategy that extracts category-level representations directly from images and corresponding masks as prompts and encourages the model to better learn the appearance differences between different categories. In this process, the model is able to better extract instance-level representations and facilitate class-agnostic segmentation. Finally,  we demonstrate for the first time that {\it category information is %still
actually 
%useful 
beneficial 
for OWIS.
}

In order to better evaluate the OWIS model, we propose a more refined and reasonable benchmark.  The existing cross-categories benchmark \cite{wang2022open,kim2022learning,hwang2021exemplar} divides the categories of the dataset into known and unknown parts (\eg, for COCO, 20 base categories  and 60 target categories ). 
However, in their setup, almost all of the unknown categories appear in the images of the training set, which can lead to %unavoidable
inevitable 
information leakage. They also ignore the fact that in the real open world, there are many classes of objects that never appear in the training set.  The cross-datasets benchmarks \cite{wang2021unidentified,wang2022open} also face the same problem. What's worse is that the improvement on those benchmarks does not even reflect stronger open-world generalization, as it may also come from %doing 
performing 
better on  known categories (due to large categories overlap between datasets).

To address the above issues, we provide a new open-world benchmark
that is 
closer to a real-world scenario by dividing the dataset classes into \textit{known-seen-unseen} parts and re-organizing the train-validation split for the COCO and LVIS datasets. See \Cref{fig:setting}. Specifically, the `seen' represents objects that appear in the training set but are not  annotated, and the `unseen' represents real-world long-tail categories that never appear in the training images.  It is worth mentioning that the existing approaches \cite{wang2022open,qi2022ssl,xue2022single} often rely on pseudo-labels, which only improve the seen categories, and no %one
work in literature 
has explored whether these methods are indeed effective on unseen categories. We are the first to emphasize the model's ability to detect unseen objects that never appear in training images.  

Experiments show that our approach \handle can improve the overall$/$unseen detection by  5.6\% and 6.1\% in AR. On existing cross-dataset transfer and strongly supervised setting, it yields 5.5\% and 12.3\% relative improvement.  We %later
show that these prompts can be generated from word embeddings and example masks, extending our methods to open vocabulary and few-shot segmentation tasks.  Our main contributions can be summarised as follows:
\begin{enumerate}
\itemsep -0.05cm 
     \item An auxiliary training mechanism, \handle, which effectively uses category information to improve class-agnostic segmentation  on various benchmarks including open-world, cross-data transfer, and strongly supervised segmentation.
     
    \item A new benchmark, \dsname, for open-world segmentation which explicitly separates ``known'', ``seen'' and ``unseen'' categories and aligns with real-world long-tail object discovery challenge. For the first time, we focus on the ``unseen'' categories that never appear in training images.   
    
    \item 
    % Our 
    %qualitative 
    Experiments demonstrate that category-level prompts do have the ability to encode corresponding category appearance representations and control mask generation, and have the potential to extend to open-vocabulary and few-shot segmentation. 
\end{enumerate}

\section{Related Work}

\begin{figure*}[htp]
    \centering
    \includegraphics[width=1\linewidth]{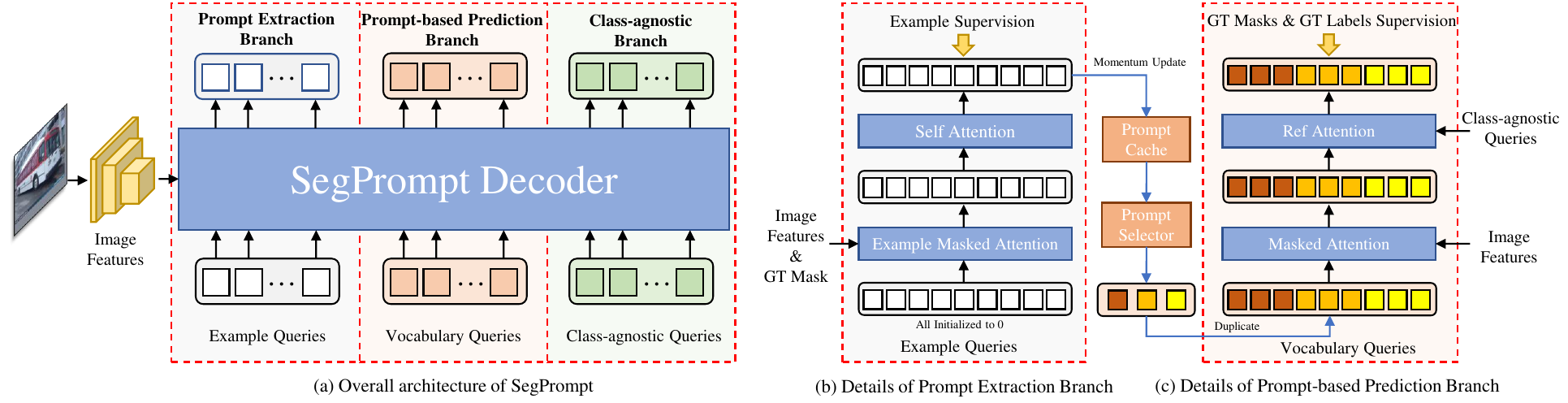}
    \caption{\textbf{The framework of \handle}. (a) 
    The  transformer decoder of our framework consists of three branches,  a class-agnostic branch, a prompt extraction branch, and a prompt-based prediction branch. All three branches share the same weight. The class-agnostic branch performs inference while the two other prompt branches are auxiliary supervision branches and can be removed during inference. Specifically, (b) the prompt extraction branch is used to perform prompt extraction and updates the prompt cache. (c) The prompt-based prediction branch performs prediction based on the prompt selected by the prompt selector as an auxiliary training module.}
    %%\vspace{-1em}
    \label{fig:framework}
\end{figure*}
\noindent \textbf{Open-vocabulary segmentation.}
Recently, %there has been 
%a lot of
many 
works %focusing
focus 
on open-vocabulary segmentation \cite{li2022language,ghiasi2022scaling,liang2022open,ding2022open,huynh2022open,xu2021simple,ding2022decoupling}. 
ZSSeg \cite{xu2021simple} establishes a simple baseline for classifying class-agnostic mask proposals via CLIP \cite{radford2021learning} image encoder and text encoder. OpenSeg \cite{ghiasi2022scaling} further exploits image-level supervision, but its category information is only input in the region word grounding part. 
In addition, XPM \cite{huynh2022open} and Ding et al.\  \cite{ding2022open} further extend the task of semantic segmentation to instance segmentation and panoptic segmentation. Existing work \cite{ghiasi2022scaling,ding2022open,huynh2022open,xu2021simple,ding2022decoupling} tends to build on the assumption that class-agnostic segmentation networks are already able to find all objects on the image, these methods usually require additional annotations, \eg, captions, for unknown classes and focus more on improving classification rather than enhancing the quality of segmentation. 
%And
They also tend to introduce too many classes in the training set, so the model focuses on semantic relations between categories and does not really reflect the ability to find unknown objects. %But
Here, 
we want to argue that discovery capability is currently the main bottleneck of the task. 
Our work, on the other hand,  is based only on limited annotations and is more concerned with how to make better use of the category information to improve class-agnostic segmentation.

\noindent \textbf{Open-world instance segmentation.} Open-world segmentation  aims to segment all objects in an image, regardless of whether their classes have been seen during training. Wang et al.\  \cite{wang2021unidentified} first proposes this task and designs the corresponding benchmark and UVO dataset. Kim  et al.\  \cite{kim2022learning} and  Qi et al.\ \cite{qi2021open} show that forcing the model to learn classification easily hampers the model's ability to generalize across categories. 

Thus, a class-agnostic segmentation network is actually more suitable for the open-world setting. 
Qi et al.\  \cite{qi2021open} propose  a new task called entity segmentation, which requires all things and stuff to be segmented out, and its model is improved based on fully-convolutional architecture. GGN \cite{wang2022open} learns pairwise affinity to generate pseudo labels, further improving the performance of the model.  
SOIS \cite{xue2022single} proposes a cross-task consistency loss to mitigate the impact of noise in the labels, and also performs semi-supervised training. 
All of the above work has removed the category information on the training sets,
%without exception, 
and most of them \cite{wang2022open,qi2022ssl,xue2022single} have used pseudo-labels for self-training. 
Our work %proves
shows 
that designing a suitable network to decouple the mask and category can effectively improve the network's generalization ability to unseen classes by simply adding limited category information.

\noindent \textbf{Text-supervised$/$unsupervised segmentation.} Training based on limited pixel-level annotations inevitably affects the ability of the model to generalize in the open world. Thus, many methods \cite{xu2022groupvit, zhou2022extract, wang2022freesolo, henaff2022object} have emerged that are not based on pixel-level annotations at all. GroupViT \cite{xu2022groupvit}, which groups semantic regions through image text comparison learning, can be successfully transferred to the task of semantic segmentation without fine-tuning. Zhou et al.\  \cite{zhou2022extract}, directly choose to use a large-scale image text pre-training model with a slight adjustment, the model is able to perform  semantic segmentation without any training.
 FreeSOLO \cite{wang2022freesolo},  successfully achieves unsupervised instance segmentation %based on 
 built upon 
 the SOLO \cite{wang2021solo} instance segmentation framework and  pre-trained features from DenseCL \cite{wang2021dense}.
 
 %after a set of self-training processes. 
% Odin \cite{henaff2022object}, on the other hand, is based on the idea of clustering and self-supervised comparative learning, which allows the model to autonomously mine the semantics in images without introducing any prior. 

However, not introducing pixel\--level label\-ing at all will un\-doubtedly lead to very poor segmentation accuracy, which is hardly %oriented
usable 
%to 
for 
practical applications in the open world. Therefore, we provide a method that can use both masks strong supervision and image-level labels for %fine-grained 
fine-granularity 
representation learning, which can guarantee the accuracy of segmentation and make full use of the information of image-level labels to improve the generalizability of the model and make the real open world segmentation possible.

\section{Methodology}
\label{sec:method}

\subsection
%\noindent
{\bf Overview} 
Open-world instance segmentation aims to segment all instances in the image, including those belonging to categories that were not present in the training phase. The existing works \cite{wang2022open,xue2022single} use a class-agnostic segmentation model to accomplish this task. In this work, we propose a prompt learning mechanism where only few prompt-related parameters are specifically supervised by categorical labels in a separated branch so that instance-level representation can be improved without affecting the class-agnostic branch.

Specifically, as shown in \Cref{fig:framework}, our prompt learning mechanism consists of a prompt extraction branch and a prompt-based prediction branch. The prompt extraction branch provides category-level instance features as prompts and the prompt-based prediction branch then uses these prompts to predict corresponding per-class instance masks. Except for few prompt-related embeddings, all parameters are shared with the class-agnostic segmentation network. 
This allows the class-agnostic branch to better model instance-level representations and only marginal overhead is added in the training. Furthermore, the whole process is auxiliary and does not affect the original inference process.

\subsection{Class-agnostic Baseline Branch}\label{subsec:clabranch}

Our auxiliary prompt learning mechanism is flexible to be applied to any transformer-based perception decoders since the prompts are only appended to the decoder queries and the model structure is kept. We choose masked attention based framework Mask2Former \cite{cheng2022masked}.
Thus the prompt features can be naturally extracted with mask guidance. The $l$-th decoder layer consists of two parts. Guided by the mask predictions from the previous layer $\mathbf{M}^{l-1}$, a masked attention module aggregates information from the image feature $\mathbf{X}$ to refine the queries $\mathbf q \in \mathbb R^{N\times D}$:
\begin{equation}
%\small
\begin{aligned}
\mathbf{q}^l_{\text{mask}}=\operatorname{MaskAttn}\left(\mathbf{q}^{l-1}, \mathbf{X}, \mathbf{M}^{l-1}\right).
\end{aligned}
\end{equation}
Then the queries are updated by a following self-attention layer:
\begin{equation}
%\small
\begin{aligned}
\mathbf{q}^l = \operatorname{SelfAttn}\left(\mathbf{q}^l_{\text{mask}}\right).
\end{aligned}
\end{equation}
Queries from the last layer $\mathbf q^L$ are used to predict the final masks and corresponding binary scores.  This branch has exactly the same supervision as the original Mask2Former except that we treat all ground-truth as one class.

\subsection{Prompt Extraction Branch}
% hc:
We maintain a prompt cache $\mathbf P = \{\mathbf p_c|c = 1, ... C\}$ for the $C$ known categories where $\mathbf{p}\in \mathbb R^D$ has the same dimension as the instance queries. To extract the corresponding prompt feature $\mathbf p_c$ for each instance from class $c$ in the training sample, a set of zero-initialized example queries $\mathbf e^0 = \{\mathbf e^0_c\}$ is added to the baseline decoder and use the output queries $\mathbf e^L$ as the extracted prompts. Different from the baseline branch, the ground-truth masks $\mathbf M_{\text{gt}}$ are used in the masked attention.
\begin{equation}
%\small
\begin{aligned}
\mathbf{e}_{\text{mask}}^{l} = \operatorname{MaskAttn}\left(\mathbf{e}^{l-1}, \mathbf{X}, \mathbf{M}_\text{gt}\right). \\
\end{aligned}
\end{equation}
We call this module \textbf{example masked attention} because it aggregates the information within a given mask and enables the model to segment unseen objects with the extracted prompts with only a few example masks. The self-attention module of the baseline branch is reused to refine the example queries,
\begin{align}
\mathbf{e}^l = \operatorname{SelfAttn}\left(\mathbf{e}_{\text{mask}}^{l}\right).
\end{align}
Notice that example queries $\mathbf e$ do not share the attention map with the baseline queries $\mathbf q$.

We also supervise each example query using mask loss, requiring the segmentation result of each query to be as consistent as possible with its corresponding  the ground-truth mask used in example masked attention, which we refer to as \textbf{example supversion}.

Finally, the example queries from the last layer $\mathbf e_c^L$ update the corresponding prompt $\mathbf{p}^c$ if they belong to the same category $c$ in a moving average fashion. For each image, we use at most 5 example queries, and we prioritize different classes:
\begin{align}
\mathbf{p}_c \gets m\mathbf{p}_c+(1-m) \mathbf e^L_c,
\end{align}
where the momentum $m$ is set to $0.9$.

\subsection{Prompt-based Prediction Branch} \label{subsec:voc}

By feeding the extracted prompts $\mathbf p^c$ to the decoder, we enforce the model to detect objects from the corresponding class $c$ in the image, namely prompt-based predictions. During training, it acts as a parallel branch and provides auxiliary class-aware supervision without affecting the main branch. 

For each training sample, a prompt selector selects and duplicates the prompts to create candidate queries $\mathbf v\in R^{C_{\text{max}}K\times D}$ for the prompt-based predictions, where $C_{\text{max}}$ is the maximum number of categories can be predicted per image and $K$ is the maximum predictions per category. If the sample contains fewer categories $C_{\text{gt}} < C_{\text{max}}$, $C_{\text{neg}} =C_{\text{max}}  - C_{\text{gt}}$ negative categories are uniformly selected from all negative classes. For each selected class $c$, the $k$-th candidate query $\mathbf v_{ck}$ is initializes as follows,
\begin{align}
    \mathbf v^0_{ck} = \mathbf p_c + \mathbf r_k + \mathbf s_c, 
\end{align}
where $\mathbf r_k\in R^D$ and $\mathbf s_c\in R^D$ are learnable embeddings. $\mathbf r$, the intra-class embedding, is shared across categories and provides an intra-class imbalance allowing queries belonging to the same category to focus on different objects, while $\mathbf s$ is class-specific embedding. 
The prompt-based prediction branch shares masked attention with the baseline branch but uses a modified self-attention module, termed reference attention $\operatorname{RefAttn}$. In reference attention, every class candidate queries $\mathbf v_{c}$ is separately encoded by the same multi-head attention layer whose weight is shared with the self-attention layer, $\operatorname{SelfAttn}$, in the baseline model. Different from the $\operatorname{SelfAttn}$, $\operatorname{RefAttn}$ not only encodes the relationship between the queries in $\mathbf v_{c}$, but also leaks the information of baseline queries $\mathbf q$ into $\mathbf v_{c}$. Specifically, $\mathbf v_{c}$ is taken as the query for the multi-head attention layer, and $[\mathbf q;\mathbf v_c]$ is taken as the key and value. The $\operatorname{RefAttn}$ can be demonstrated by:
%
%%%\vspace{-0.2em}
\begin{equation}
\centering
\begin{aligned}
\mathbf v^{updated} &= \operatorname{RefAttn}\left(\mathbf v\right) 
                    = \operatorname{Concat}\left(\mathbf v_0^{\prime}, ..., \mathbf v_{C_{max}}^{\prime}\right), \\
\text{where}& \ \mathbf v_c^{\prime} = \operatorname{MultiHead}\left(\mathbf v_c, [\mathbf q;\mathbf v_c], [\mathbf q;\mathbf v_c]\right) \\ 
\end{aligned}
\end{equation}
where $\mathbf v^{updated}$ are the candidate quires updated by $\operatorname{RefAttn}$. By doing so, the candidate queries can fully utilize the information of the class-agnostic branch. Meanwhile, the candidate queries are invisible to the candidate queries whose class is different, which is to prevent the model from learning trivial solutions, such as some categories always appearing in pairs, and at the same time avoiding unnecessary interference from negative sample categories.

For the loss calculation, we adopt the intra-class many-to-one matching strategy. For queries corresponding to a category $\mathbf{c}$, if $\mathbf{c}$ is a negative sample, only the classification loss is calculated and it is not involved in the matching, and its ground-truth label is set to 0. If $\mathbf{c}$ is a positive sample, those queries are matched to the ground-truth  instances whose class are $\mathbf{c}$. After the matching, the ground-truth  labels of matched queries are set to $1$ and the mask loss is calculated with the matched ground-truth  mask.
The ground-truth  labels of unmatched queries are set to $0$, and the mask loss is not calculated.

\section{Benchmark}

\subsection{Task Formulation}

Open-world instance segmentation is to segment all the object instances of any class, regardless of whether the class appears in the training phase  \cite{wang2021unidentified,qi2021open,xue2022single}. 
Formally, in open-world instance segmentation, the training set $\mathcal{D}_{train}$ contains various classes of densely annotated objects, denoted as $C_{train}$. 
Models trained on $\mathcal{D}_{train}$ are supposed to segment all the objects that appear in the images of the test set $\mathcal{D}_{test}$, which contains a set of classes $\mathcal{C}_{test}$. 
Generally, we have $\mathcal C_{train} \subset \mathcal C_{test}$.

Previous work \cite{wang2022open,kim2022learning} divides the categories of the dataset into known and unknown. 
As mentioned in \cref{sec:intro}, by splitting a specific dataset or using a cross-dataset setting, they guarantee $ C_{train} =  C_{known}$ and $ C_{test} =  C_{known} + C_{unknown}$. 
However, almost all of the unknown categories appear in the images of the training set, which leads to information leakage. 

To address this issue, we divide the object categories into three parts: `known' $\mathcal{C}_{known}=\left\{c_k\right\}_{k=1}^{K}$, `seen' $\mathcal{C}_{seen} = \left\{c_s\right\}_{s=1}^{S}$ and `unseen' $\mathcal{C}_{unseen}= \left\{c_{u}\right\}_{u=1}^{U}$. 

The known objects in $\mathcal{C}_{known} $ are those with dense annotations and category information in the training set $\mathcal{D}^{train}$, while the objects in $\mathcal{C}_{seen}$ appears on the image of the training set but are not been annotated. 
By contrast, the unseen objects do not appear in the training set at all. 
Thus, we have $ C_{train} = C_{known}$ and $ \mathcal{C}_{test} = \mathcal{C}_{known} \cup \mathcal{C}_{seen} \cup \mathcal{C}_{unseen}$.

\subsection{Dataset Construction}

\noindent\textbf{Base datasets.}
We construct our dataset based on the COCO dataset \cite{lin2014microsoft} and the LVIS dataset \cite{gupta2019lvis}. 
The COCO dataset is a widely used large-scale instance segmentation benchmark.
COCO contains $118$k images for training, $5$k images for validation, and $41$k images for testing. 
$80$ categories of objects are annotated on the dataset.
LVIS is recently proposed for large vocabulary instance segmentation.
Sharing the same images with COCO, LVIS has $1203$ categories.
LVIS re-splits the images of COCO and contains $100$k/$20$k images for training/validation, respectively.
The categories of LVIS are divided into three groups: frequent, common, rare, based on the number of training images. 

\begin{table}[t]
\begin{center}
\scalebox{0.86}
{
\begin{tabular}{c | c c |  c c c}
\toprule
\multirow{2}{*}{} & \multicolumn{2}{c|}{Train} & \multicolumn{3}{c}{Test} \\
& Known & Seen & Known & Seen & Unseen \\
\midrule
$N^{cls}$ & 123 & 743 & 123 & 738 & 337 \\
$N^{ins}$ & 1,511,257 & 452,524  & 168,798 & 99,230 & 6,457 \\
\bottomrule
\end{tabular}
}
\end{center}
\vspace{-1em}
\caption{\textbf{The statistics of the proposed dataset}. The $N^{cls}$ and the $N^{ins}$ are the number of classes and instances.}
\vspace{-1em}
\label{tab:dataset-sta}
\end{table}

\noindent\textbf{Data organization.}
Previous work \cite{wang2022open,kim2022learning} based on both COCO and LVIS uses the images of the training/validation split of LVIS as the training/testing set, but uses annotations of COCO/LVIS for the training/testing images. 
To divide the unknown classes into the seen and unseen, we further remove the images containing objects of unseen classes of the training set and added them to the testing set. 
In the following, we illustrate how we determine the three groups of classes.

For the $\mathcal{C}_{known}$, we first choose the $80$ categories of COCO because of the high quality of annotations of COCO. 
Considering there are obvious annotation granularity differences between COCO and LVIS, we add annotations of some frequent categories in LVIS to enable the model to segment objects at a finer granularity.
Specifically, for an object with the label ``person'' in COCO, some parts of it will be labeled as  ``apron'', ``hat'', etc., in LVIS.
If only COCO is used for training, the model can locate a whole person, but can not segment the apron or hat of the person, which makes the OWIS task intractable. 
Thus we select the most frequent $64$ categories of LVIS and the $80$ categories of COCO to compose the known classes $\mathcal{C}_{known}$. 

For the $\mathcal{C}_{unseen}$,  
we conduct experiments to explore the impact of removing images from the training set, and select the $337$ classes of objects, which belongs to the rare split of LVIS, as the unseen classes $\mathcal{C}_{unseen}$. 
Only less than $1\%$ images of the training set are removed. 
The details can be found in the supplementary material. 
The other classes (except the known and unseen classes) of LVIS are regarded as seen classes $\mathcal{C}_{seen}$ of our dataset. 

\noindent\textbf{Dataset analyses.}
After the re-splitting of COCO and LVIS, the training/testing set of the \dsname dataset contains $98708$/$21271$ images.    
\Cref{tab:dataset-sta} shows detailed statistics.
Note that the known subset contains $123$ rather than $144$ classes due to the overlap between COCO and LVIS.
Besides, some classes belonging to the seen subset of the training set do not appear in the test set. 
But this does not affect the model training and evaluation.

\subsection{Metrics}
\label{subsec:metrics} 
To fully evaluate the ability of models to segment unseen objects, we compute the class-wise mean Average Recall (AR) on different sets as $AR_{known}$, $AR_{seen}$,  $AR_{unseen}$, and $AR_{all}$.
For a particular class, we use all class-agnostic predictions and ground-truth instances belonging to the class to compute the AR, as \cite{saito2021learning}. 
Subsequently, the ARs of classes of a specific subset are then averaged to obtain the AR on the subset. 
Considering the number of instances in each subset is of different orders of magnitude, this division also helps to alleviate the impact of class imbalance. 

We %didn't
do not 
use the class-agnostic AR and the mean Average Precision (AP) as previous work \cite{wang2022open,xue2022single}, because the two metrics are not suitable to evaluate open-world models on datasets with a large vocabulary such as LVIS.
Firstly, AP is unable to properly measure the ability of models to discover unseen classes.
Objects that are not in the test set annotation but found by the model will be counted as false positives, and result in a lower AP. 
% , as revealed by \cite{bansal2018zero}.
Secondly, class-agnostic AR will be dominated by common classes with more instances, and thus can not evaluate the ability of models on long-tailed categories. For simplicity, the class-wise mean $AR_{cls}$ is referred to as $AR$ in later sections. 

\section{Experiments}

\subsection
%\noindent 
{\bf Implementation Details}
All experiments are conducted on eight Nvidia A100 GPUs. ResNet50 is used as the backbone for all models. Most of experimental settings follow Mask2Former \cite{cheng2022masked}, including the learning rate, weight decay, data augmentation and batchsize. Unless specified, the number of object queries for the class-agnostic branch is 100, while the total number of object queries for the prompt-based prediction branch is 300.

\begin{table}[b!]
\center
\resizebox{1\columnwidth}{!}{%
\begin{tabular}{ l | c | c c c | c  c  c}
\toprule
Method               & $AR_{all}$    & $AR_{kn}$ & $AR_{se}$ & $AR_{un}$& $AR_{s}$      & $AR_{m}$      & $AR_{l}$       \\ \midrule
Mask R-CNN              & 36.5          & 46.5          & 32.8          & 41.0                  & 24.2          & 44.3          & 57.0                    \\
Mask R-CNN$^\dag$      & 39.9          & 46.6          & 36.8          & 44.0                  & 28.3          & 49.3          & 57.3                    \\
Mask2Former             & 40.8          & 50.2          & 37.4          & 44.9                  & 25.5          & 50.0          & 67.9                    \\
Mask2Former$^\dag$     & 42.5          & 49.9          & 39.0          & 47.4                  & 27.2          & 51.5          & 68.1                    \\
LDET  \cite{saito2021learning}           & 39.3           & 40.7         & 37.6        & 42.6    & 27.4     & 49.2     & 55.7       \\
GGN  \cite{wang2022open}                 & 42.3           & 42.6         & 40.3        & 46.6          & 28.3     & 52.9     & 62.1  \\
Ours                 & \textbf{44.9}  & \textbf{52.2} & \textbf{41.2} & \textbf{50.3} & \textbf{29.4} & \textbf{54.5} & \textbf{70.4}  \\
\bottomrule
\end{tabular}%
}
%%\vspace{-0.5em}
\caption{\textbf{Class-wise mean AR comparison of different methods.} $^\dag$ denotes class-agnostic training. Our prompt learning mechanism improves the performance on known and unseen categories by 2.3 AR and 2.6 AR for free.}
 %%\vspace{-1em}
\label{tab:benchmark}
\end{table}

\subsection{Main Results}
\noindent \textbf{Experimental results on our benchmark.} In \Cref{tab:benchmark}, we compare our method with the two most representative strong baselines Mask R-CNN \cite{he2017mask} and Mask2Former \cite{cheng2022masked} on our benchmark. Notably, Mask2Former outperforms Mask R-CNN in the vast majority of metrics but is significantly weaker than Mask R-CNN in $AR_{s}$ instead. We also demonstrate that class-agnostic training does significantly improve generalization to unknown categories, with about 2\% to 4\% improvements in both $AR_{seen}$ and $AR_{unseen}$, but for known categories, the difference is actually very small, and
 can be considered as a fluctuation. In contrast, our method further improves generalizability by effectively using the category information instead, with very significant improvements in all metrics, especially $AR_{seen}$ and $AR_{unknown}$. It is worth mentioning that much of the work \cite{ding2022decoupling,huynh2022open,xu2021simple,ding2022open} can actually be attributed to the class-agnostic Mask2Former \cite{cheng2022masked} model in \Cref{tab:benchmark} or a weaker network structure. Our experiments show that, even under the same amount of data, there is still a lot of room for improvement in the performance of Mask2Former. How to use limited annotations to improve the performance of the model is still a topic worth studying.

\noindent  \textbf{Evaluation on other benchmark.}
To verify the effectiveness and universality of our method, we perform experiments following the cross-dataset evaluation fashion proposed by the work \cite{xue2022single,saito2021learning}, as shown in \Cref{tab:coco-lvis}. All the models are trained on COCO dataset but evaluated in LVIS dataset, denote as COCO$\rightarrow$ LVIS. The experiment results show that our method outperforms the current the-state-of-the-art method SOIS \cite{xue2022single} in this setting. Both SOIS and our method are based on Mask2Former \cite{cheng2022masked} methods, but SOIS also performs pseudo labeling and self-training methods, while our method does not need to perform self-training so if the number of
training iterations is the same for both rounds, our training
overhead is only half. We also show the results of whether to introduce example supervision or not, as we find that while introducing example supervision leads to an overall performance improvement, $AP_{l}$decreases. This is consistent with the results in \Cref{tab:introcat}, where we speculate that example supervision allows the model to focus better on small objects, but not on large objects.

\noindent \textbf{Training on the full LVIS training set.}
We further validate the effectiveness of our method when the fully supervised training on the full LVIS dataset is performed, as shown in \Cref{tab:fulllvis}. Compared to Mask2Former, Our method achieves a leap in all metrics when all the setting is strictly aligned, including the training data, the number of training iterations and so on. This shows that our method has a very obvious advantage in performing class-agnostic segmentation training for such a complex scenario in terms of object types and numbers.

\begin{figure*}[t!]
    \centering
    \includegraphics[width=0.987\linewidth]{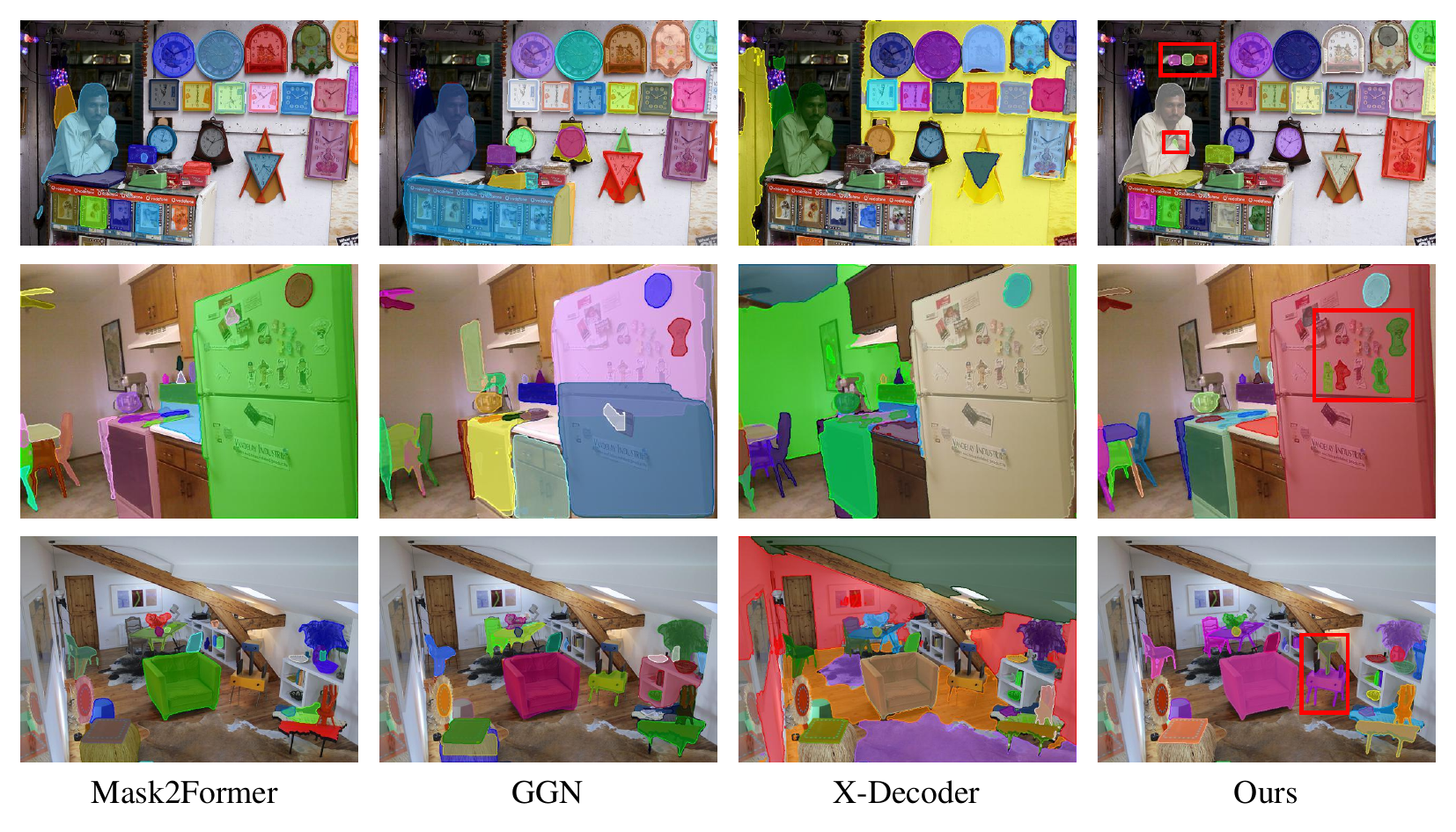}
    %%\vspace{-1em}
    \caption{\textbf{Qualitative results on OWIS}. We compare our method with Mask2Former \cite{cheng2022masked}, GGN \cite{wang2022open}, and X-Decoder \cite{zou2022generalized} (X-Decoder is an open-vocabulary model, so we add LVIS class texts as guidance) and select top 30 masks for visualization. The additional objects that were detected by our method are highlighted with red boxes. All models except X-Decoder are trained using COCO annotation only.}
    %%\vspace{-1em}
    \label{fig:open}
\end{figure*}

\begin{table}[]
\resizebox{\columnwidth}{!}{%
\begin{tabular}{l | c | c | c  c  c | c}
\toprule
Method         & $AP$@100  & $AR$@100 & $AR_{s}$ & $AR_{m}$ & $AR_{l}$ & $AR_{all}$  \\ \midrule
Mask2Former$^\dag$ & 20.3          & 40.9                           & 26.2                         & 61.5                         & 78.5                         & 50.8        \\
Mask2Former          & 20.1          & 41.7                           & 26.4                         & 63.0                           & 80.8                         & 53.8        \\

Ours                 & \textbf{22.8} & \textbf{45.7}                  & \textbf{30.2}                & \textbf{68}                  & \textbf{84.2}                & \textbf{57.4} \\
\bottomrule
\end{tabular}%
}
%%\vspace{-0.5em}
\caption{\textbf{Comparison on standard LVIS dataset.} $^\dag$ denotes class-agnostic training. Except for $AR_{all}$, which is our metric, all other metrics are from the original LVIS class-agnostic evaluation. Our method significantly improves all the metrics by at least 2\%.} 
%%\vspace{-1em}
\label{tab:fulllvis}
\end{table}

\begin{table}[h]
\center
\resizebox{0.980\columnwidth}{!}{
\begin{tabular}{l | c | c | c  c  c }
\toprule
Method           & $AR_{100}$    & $AP_{100}$   & $AP_{s}$   & $AP_{m}$      & $AP_{l}$    \\ \midrule
Mask R-CNN \cite{xue2022single}    & 22.4          & 6.5          & 3.2          & 10.4          & 17.6          \\
LDET \cite{xue2022single}        & 25.1          & 6.5          & 2.8          & 10.6          & 18.4          \\
SOLO2 \cite{xue2022single}     & 21.7          & 7.5          & 3.1          & 12.4          & 22.4          \\
SOLO2  +   SOIS \cite{xue2022single}  & 22.8          & 7.9          & 3.8          & 13.1          & 23.8          \\
M2F  \ \ \ \ \   +   SOIS   \cite{xue2022single}      & 25.2          & 8.5          & 3.4          & 13.8          & 26.4          \\

GGN \cite{wang2022open}              & 27            & 6.5          & 3.5          & 12.1          & 14.9    \\ \midrule
Ours$^{\star}$            & 26.5          & 9.3          & 4.0          & \textbf{15.1} & \textbf{28.0} \\
Ours           & \textbf{28.5} & \textbf{9.4} & \textbf{4.6} & \textbf{15.1} & 24.5    \\
\bottomrule
\end{tabular}
}
%%\vspace{-0.5em}
\caption{\textbf{Comparison on cross-dataset evaluation setting, COCO$\rightarrow$LVIS.} The models are trained on COCO dataset but evaluated on LVIS dataset. The original LVIS evaluation metric is employed. Our method performs better over all metrics. For GGN, the original paper did not report the results under this setting, so we used their model trained under COCO to re-evaluate. $\star$, means we do not use example supervision.
}

 %%\vspace{-1em}
\label{tab:coco-lvis}
\end{table}

\subsection{Ablation Study}
In the ablation study, we carefully investigate the design of each component we proposed. And the number of the maximum prediction per category $K$ is set to 10 and the maximum number of categories $C_{\text{max}}$ that can be predicted per image is set to 30, if not specified. More ablation experiments are available in the supplementary material.

\noindent\textbf{Auxiliary supervision methods.}
The proposed prompt learning is used as an auxiliary supervision to supervise the model with label information. However, there are many different auxiliary supervision methods to introduce the category information to the model. Here, we explore a straightforward auxiliary supervision method to compare with \handle, as shown in \Cref{tab:introcat}. The straightforward method is that a vanilla Mask2Former is used in the auxiliary supervision branch for classification, similar to the prompt-based prediction branch, while the weights of the vanilla Mask2Former are shared by the class-agnostic baseline branch except for the classification head. 

The main difference between this straightforward method and \handle is that each query in this method does not specify a corresponding category, but matches with the ground-truth of all categories directly in the matching phase for multi-category prediction. The experiment results demonstrate that our method outperforms the vanilla mask2Former significantly and the prompt learning is the key to take advantage of category information without losing unknown class generalizability. In \Cref{tab:introcat},  we also compare the impact of introducing example supervision or not, and the results show that adding example supervision can further improve the model performance.

\begin{table}[t]
\center
\resizebox{\columnwidth}{!}{%
\begin{tabular}{l | c | c  c  c | c  c  c }
\toprule
Aux. Sup.           & $AR_{all}$    & $AR_{kn}$  & $AR_{se}$   & $AR_{un}$ & $AR_{s}$          & $AR_{m}$          & $AR_{l}$\\ \midrule
None                        & 42.5          & 49.9          & 39.0          & 47.4          & 27.2              & 51.5              & 68.1    \\
Vanilla M2F         & 43.0          & 51.4          & 39.3          & 47.9          & 27.6              & 52.0              & 69.7\\
Ours$^{\star}$    & 44.1         & 51.9          & 40.6          & 49.1       & 28.9          & 53.3          & \textbf{70.6}      \\
Ours & \textbf{44.9}    & \textbf{52.2} & \textbf{41.2} & \textbf{50.3} & \textbf{29.4} & \textbf{54.5} & 70.4   \\

\bottomrule
\end{tabular}%
}
%%\vspace{-0.5em}
\caption{\textbf{Investigating the auxiliary supervision methods.} For `None', the auxiliary supervision is removed from the model. For `Vanilla M2F', a standard Mask2Former is used as an auxiliary supervision and shares weights with the baseline model. $\star$ means that we do not use example supervision.}
%%\vspace{-1em}
\label{tab:introcat}
\end{table}

\noindent\textbf{The way to generate the candidate queries.}  
As mentioned in \Cref{subsec:voc}, both the prompt $\mathbf{p}$ and class-specific embedding $\mathbf s$ are all the important components to generate the candidate queries $\mathbf v$. Here, we conduct experiments to verify the necessity of the class-specific embedding $\mathbf s$ and the prompt $\mathbf{p}$ by removing them from the candidate queries $\mathbf v$, separately.

As shown in \Cref{tab:prompt}, without any of the class-specific embedding $\mathbf s$ or the prompt $\mathbf{p}$, the performance will drop significantly which indicates both of $\mathbf s$ and $\mathbf{p}$ are necessary for our model.

\begin{table}[h]
\center
\resizebox{\columnwidth}{!}{%
\begin{tabular}{l | c | c  c  c | c  c  c}
\toprule
Method                      & $AR_{all}$        & $AR_{kn}$      & $AR_{se}$       & $AR_{un}$     & $AR_{s}$          & $AR_{m}$      & $AR_{l}$ \\ \midrule
Baseline                    & \textbf{44.1}     & \textbf{51.9}     & \textbf{40.6}     & \textbf{49.1}     & \textbf{28.9}     & \textbf{53.3} & \textbf{70.6}     \\ 
Baseline w/o $\mathbf s$    & 43.3              & \textbf{51.9}     & 39.6              & 48.3              & 28.0              & 52.7          & 70.1     \\
Baseline w/o $\mathbf p$    & 42.8              & 50.8              & 39.2              & 47.7              & 27.5              & 52.0          & 69.0     \\ \bottomrule
\end{tabular}%
}

%%\vspace{-0.5em}
\caption{\textbf{Verifying the necessity of the class-specific embedding $\mathbf s$ and the prompt $\mathbf{p}$.} `Baseline' represents our model without example supervision. `w/o $\mathbf s$' and `w/o $\mathbf p$' represent removing the class-specific embedding $\mathbf s$ and $\mathbf p$ from the candidate query $\mathbf v$.
%, respectively.
}
%%\vspace{-1em}
\label{tab:prompt}
\end{table}

\begin{table}[h]
\resizebox{\columnwidth}{!}{%
\begin{tabular}{l | c | c  c  c | c  c  c}
\toprule
Query Attention                                                         & $AR_{all}$ & $AR_{kn}$ & $AR_{sn}$ & $AR_{un}$ & $AR_{s}$ & $AR_{m}$ & $AR_{l}$ \\ \midrule
{$\operatorname{SelfAttn}$}                        & 35.7       & 44.1      & 31.6      & 41.4      & 22.6     & 43.2     & 58.5     \\
$\operatorname{SelfAttn}$ (independent)                                  & 41.0       & 47.5      & 37.3      & 46.5      & 26.0     & 49.6     & 67.8     \\
{$\operatorname{RefAttn}$ (inter-class visible)}  & 44.0       & 51.8      & 40.5      & 48.9      & 28.2     & \textbf{53.7}     & 70.1     \\
{$\operatorname{RefAttn}$ (inter-class invisible)} & \textbf{44.1}      & \textbf{51.9}      &\textbf{40.6}     &\textbf{49.1}      & \textbf{28.9}    & 53.3     & \textbf{70.6}     \\ \bottomrule
\end{tabular}%
}
%%\vspace{-0.5em}
\caption{\textbf{Comparison of different attention designs}. $\operatorname{SelfAttn}$ (independent) represents the prompt-based prediction branch and class-agnostic baseline branch conduct independent self-attention, and RefAttn (inter-class visible/invisible) represents whether to make queries between classes visible when reference-attention is performed.}
 %%\vspace{-0.5em}
\label{tab:selfatt}
\end{table}
\noindent \textbf{Reference attention.} Experiments in \Cref{tab:selfatt} are conducted to verify the effectiveness of reference attention $\operatorname{RefAttn}$.  
In the experiments, we compare the proposed approach with three other designs. First, we use the global self-attention  making the class-agnostic query and vocabulary query fully visible to each other. This definitely causes information leakage and impairs the segmentation ability of category-independent branches. Second, we let the two branches perform completely independent self-attention, which also does not facilitate class-agnostic segmentation. Next, we explore the impact
of visibility between different classes of candidate queries when doing the reference attention and find no significant
difference.

\begin{figure}[tp]
    \centering
    \resizebox{0.8\columnwidth}{!}{
    \includegraphics[width=0.99\linewidth]{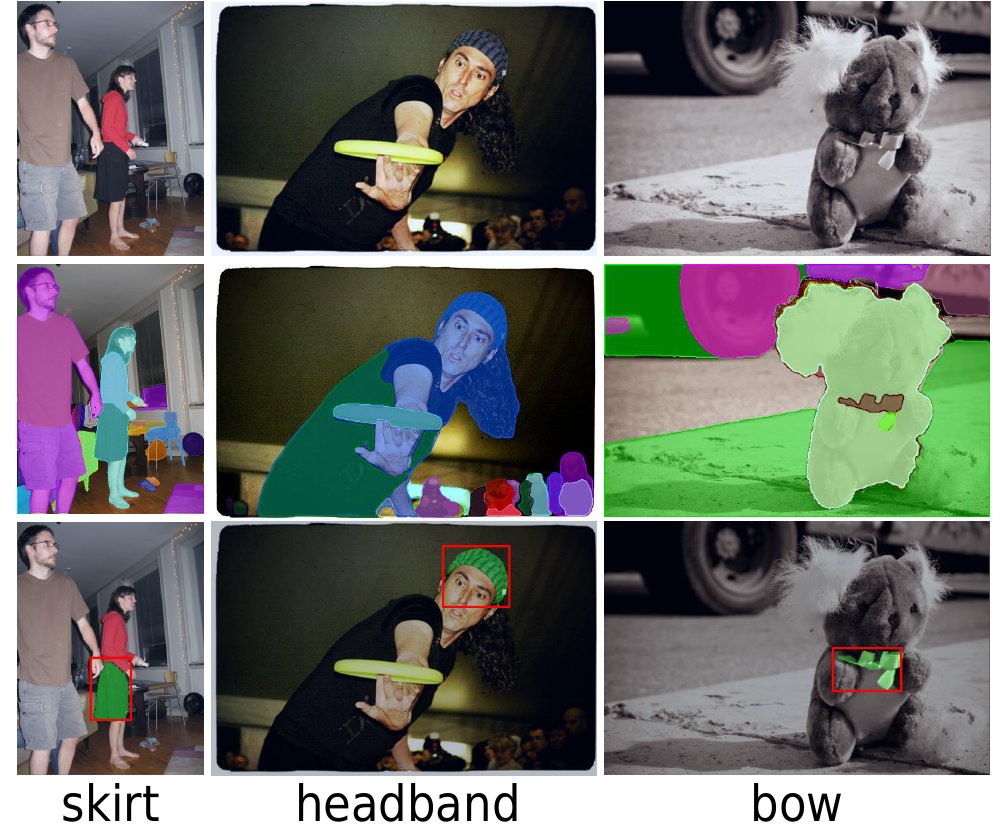}
    }
    \caption{\textbf{Qualitative results on open vocabulary segmentation.} The first row is the original input image; the second row is the result of the class-agnostic branch segmentation; the third row is the result of using CLIP text embedding as a prompt.  }
    %%\vspace{-1.5em}
    \label{fig:openvoc}
\end{figure}

\begin{figure}[h]
    \centering
    \includegraphics[width=1\linewidth]{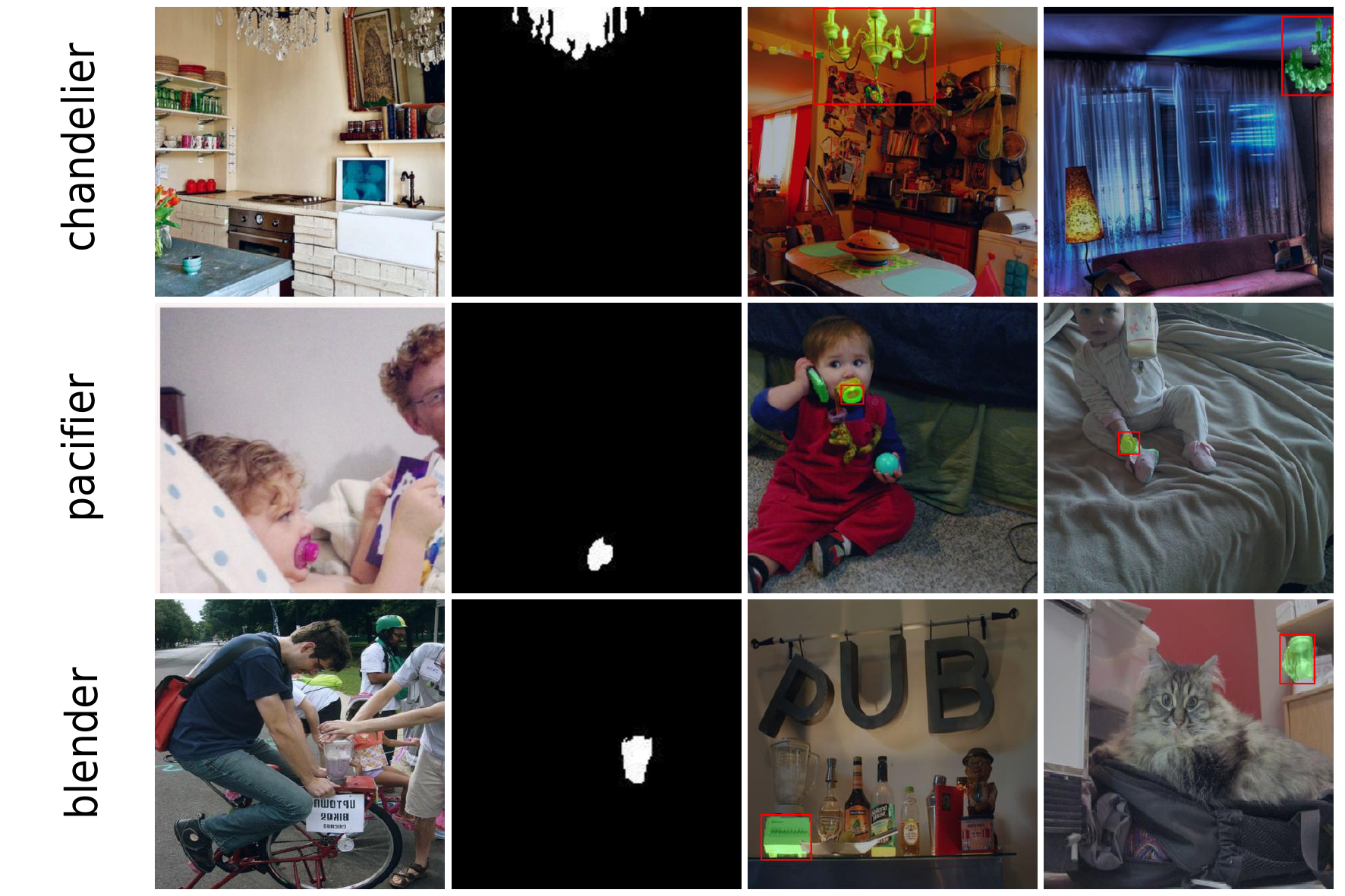}
    \caption{\textbf{Qualitative results on few-shot segmentation}. The prompts on the left side are the support images and their corresponding masks (we use the 5-shot setting, only one is shown here), followed by the segmentation results on test images.}
    %%\vspace{-1.5em}
    \label{fig:few-shot}
\end{figure}

\subsection{Qualitative Results}
\label{subsec:app}
We first compare our model with other methods on OWIS, see \Cref{fig:open}. 
To verify that our framework is indeed able to encode category-level appearance representation in the prompt, we show some qualitative results  when the setting is extended to few-shot segmentation \cite{shaban2017one,min2021hypercorrelation,liu2023matcher} and open-vocabulary segmentation \cite{li2022language,ghiasi2022scaling}. 

\noindent \textbf{Open vocabulary segmentation.}  Our framework is able to easily extend to the open-vocabulary segmentation task without changing the training process. We set the prompt of  the prompt-based prediction branch to fixed CLIP \cite{radford2021learning} text embedding, without introducing the image feature of CLIP. By adding the image-level label of the seen part during training, our model is able to segment the new classes specified by the text without any pixel annotation, see Figure \ref{fig:openvoc}.

\noindent \textbf{Few-shot instance segmentation.} Our framework can also be naturally transferred to the few-shot segmentation task. With the help of the prompt extraction branch, we can input a limited number of new classes of masks and images to the prompt extraction branch to extract the prompt of the corresponding class. The prompt is then used to inference on other images containing this new class to obtain the mask. The visualization results are shown in Figure \ref{fig:few-shot}.

\section{Conclusion}
In this work, we have presented an \dsname benchmark for open-world instance segmentation.
We introduced \textbf{\handle}, a novel prompt learning mechanism, for OWIS.
It enables segmentation models to predict masks specified by the prompt and significantly improves the class-agnostic segmentation capability of the model.
Extensive experiments demonstrated that the proposed \handle can effectively discover objects of categories beyond the training phase.

\subsection*{Acknowledgements} 
This work was supported by the National Key R\&D
Program of China (No.\  2022\-ZD\-0\-1\-1\-8\-7\-0\-0),

{\small
%\normalem
\bibliographystyle{ieee_fullname}
\bibliography{ICCV}
}

\input _supp.tex

\end{document}

%% file: _constants.tex
\def\eg{{e.g.\xspace}}

\def\handle{{SegPrompt}\xspace}

\def\dsname{{LVIS-OW}\xspace}

\def\paperTitle{\handle: Boosting Open-World Segmentation \\ via  Category-level Prompt Learning}
\newif\ifreview 
\newif\ifarxiv 
\newif\ifcamera 
\newif\ifrebuttal 

%% file: _supp.tex
\clearpage 
\appendix

\section{Dataset Analysis}
\subsection{Training set}
In this section, we analyze the differences in training annotations between COCO   \cite{lin2014microsoft}, LVIS   \cite{gupta2019lvis} and our benchmark LVIS-OW, and verify the rationality and necessity of our constructed training set. 

\begin{figure*}[tp]
    \centering
    \includegraphics[width=0.9\linewidth]{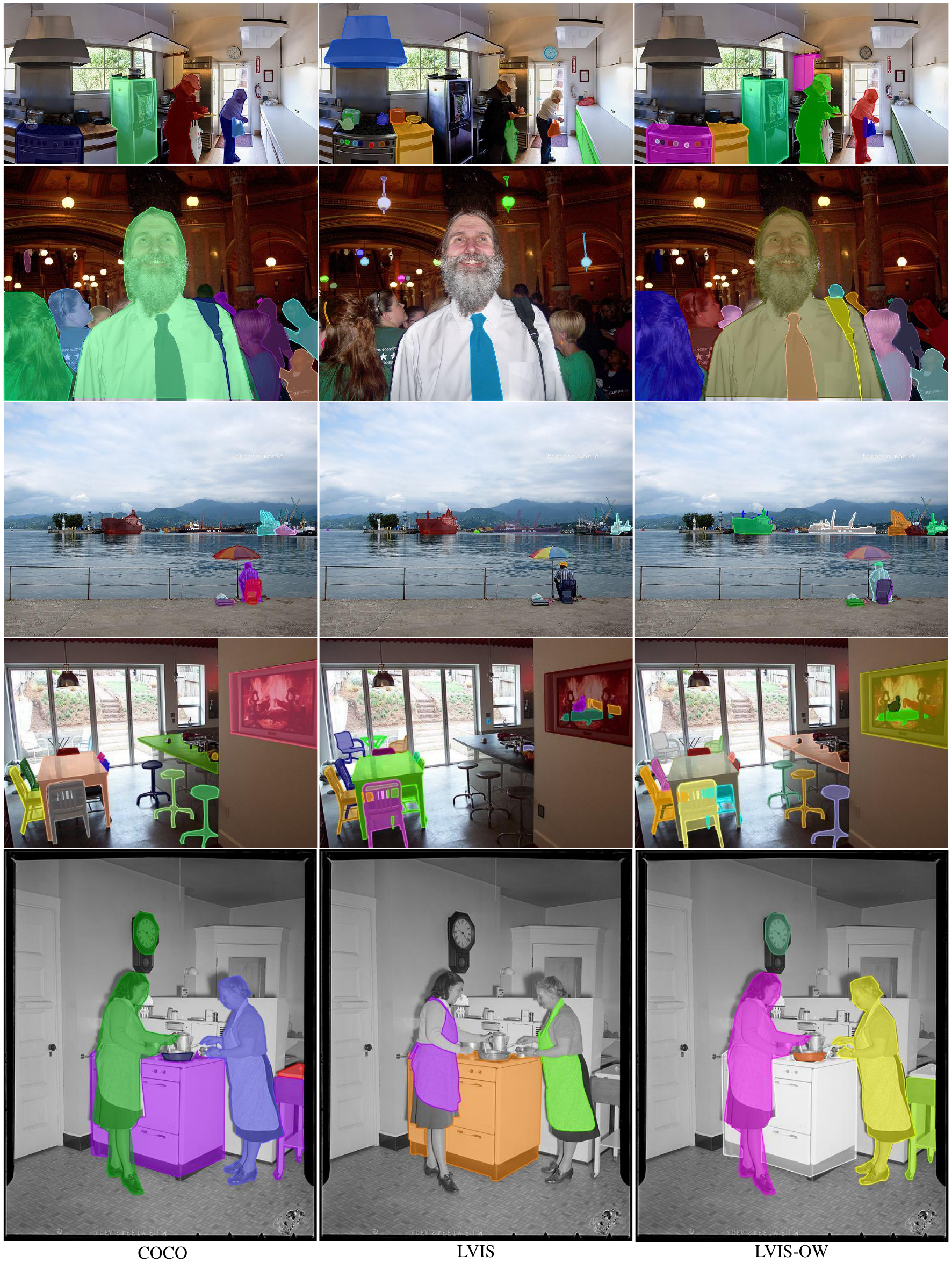}
    \caption{\textbf{Comparison of COCO, LVIS and our dataset.} COCO's annotation is more elaborate, but at a coarser granularity, and tends to annotate only instances of ``person''. In contrast, LVIS's annotation is very sparse but will focus on more detailed objects, for example, it tends not to segment ``person'', but will segment objects such as ``apron'', ``tie'', and ``hat'' on ``person''.Our benchmark combines the advantages of both and is more oriented towards an open-world setting.}
    \label{fig:dataset_vis}
\end{figure*}

As shown in  \Cref{fig:dataset_vis}, compared with LVIS' non-exhaustive annotations, COCO's annotations miss fewer objects which are in $\mathcal{C}_{known}$, and consequently reduce the ambiguity caused by annotations, so it is more suitable for training. However, there are obvious  granularity
differences between COCO  and LVIS, for example, COCO only labels the complete ``person'', while LVIS labels the items on the person such as ``skirt'', ``shoes'', etc.  \Cref{tab:worst} shows that if we use COCO exclusively for training, the model is completely unable to segment many common categories of LVIS, which is difficult to solve at the model and method levels. Our  benchmark can alleviate this problem very well by introducing a small number of categories (64 categories, about 50\% of all instances). At the same time, we  remove some annotations of rare objects, making our dataset more suitable for open-world evaluation.
\begin{table}[h]
    \centering 
\resizebox{.8\columnwidth}{!}{%
\begin{tabular}{@{}lcccc@{}}
\toprule
name               & \multicolumn{1}{c}{\#instance} & \multicolumn{1}{c}{LVIS-OW} & \multicolumn{1}{c}{COCO} \\ \midrule
cupboard           & 329                           & 53.2                             & 3.4                             \\
polo shirt         & 371                           & 54.5                             & 9.5                             \\
sweatshirt         & 258                           & 60.2                             & 16.5                            \\
tank top(clothing) & 337                           & 43.0                             & 4.1                             \\
billboard          & 270                           & 43.2                             & 6.3                             \\
jean               & 971                           & 38.6                             & 6.2                             \\
brake light        & 210                           & 30.6                             & 2.4                             \\
blinker            & 238                           & 26.7                             & 1.7                             \\ \bottomrule
\end{tabular}%
}

\caption{\textbf{The results of models trained with different training sets on several categories of LVIS.} The LVIS-OW and COCO represent training on different training sets, and the evaluation metric is class-wise AR in Sec. 4.3 in the main text. We selected classes with a high number of occurrences on LVIS validation set (\#instance \textgreater{} 100) and a large AR gap. Many classes are completely undiscoverable relying only on COCO training.}
\label{tab:worst}
\end{table}
 
\begin{table}[ht]
\centering
    \resizebox{.8\columnwidth}{!}{%
\begin{tabular}{@{}cccc@{}}
\toprule
removal ratio & \#image & \#class & \# image per class \\ \midrule
1\%(R)        & 1472    & 337     & \textless{}10      \\
5\%           & 5000+   & 544     & \textless{}29      \\
10\%          & 10000+  & 688     & \textless{}55      \\
20\%          & 20000+  & 846     & \textless{}134     \\ \bottomrule
\end{tabular}%
}
\caption{\textbf{Detailed information on the different removal ratios.} We also show the total number of images, the total number of instances to be removed  and the frequency of images for each class. The removed images are all added to the original test set lvis\_v1\_val, and all the removed categories are treated as unseen set $\mathcal{C}_{unseen}$.}
\label{tab:unseensplit}
\end{table}

\begin{figure*}[tp]
    \centering
    \includegraphics[width=1\linewidth]{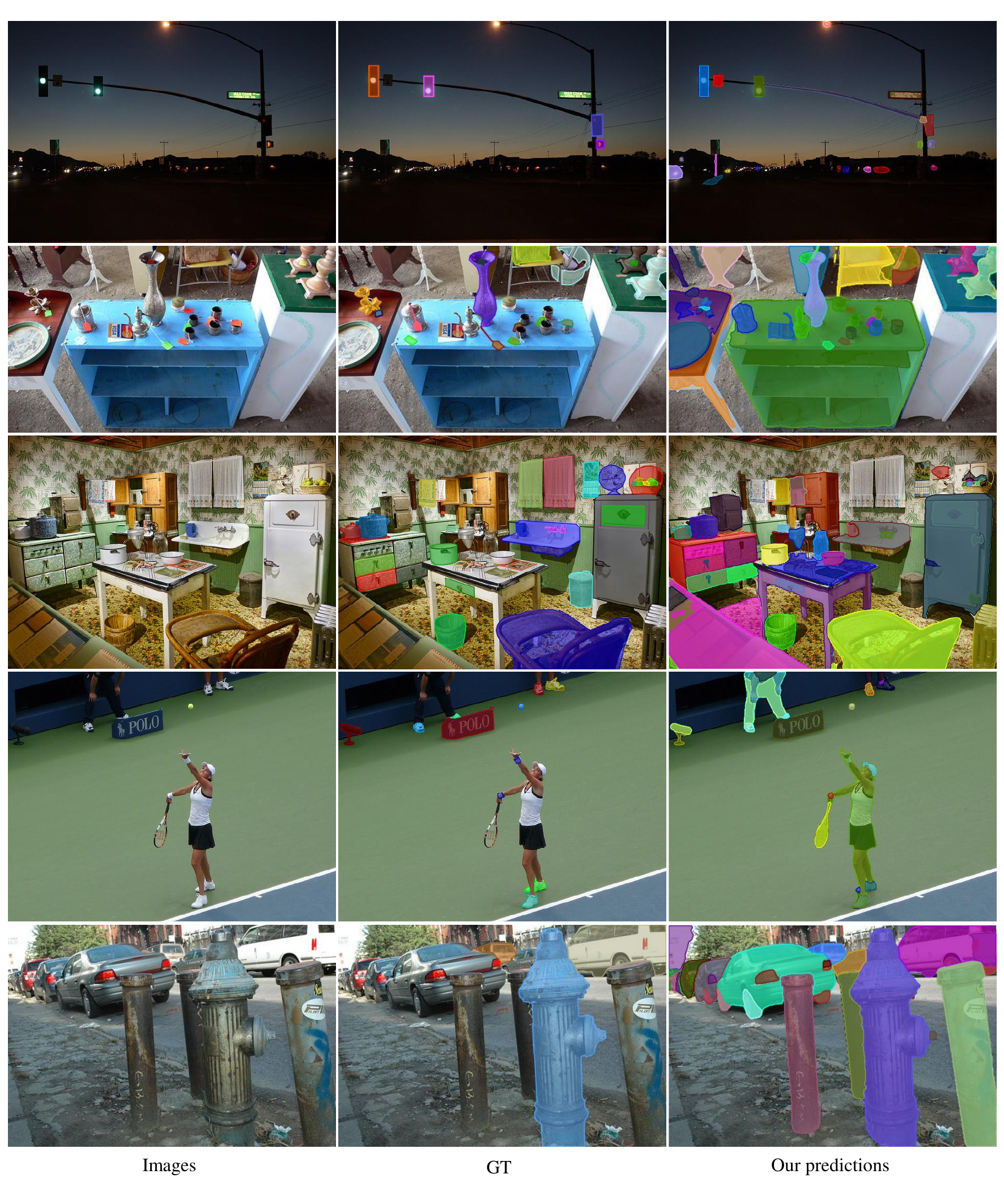}
    \caption{\textbf{Qualitative comparison of ground truth  and our prediction} Our method can detect many objects that are not in the ground truth, which includes both objects belonging to unseen set $\mathcal{C}_{unseen}$ and objects that are in $\mathcal{C}_{known}$ or $\mathcal{C}_{seen}$ but are missed by the annotations.}
    \label{fig:gtandours}
\end{figure*}

% \usepackage{booktabs}
% \usepackage{multirow}
% \usepackage{graphicx}
% \usepackage[normalem]{ulem}

% \usepackage{booktabs}
% \usepackage{multirow}
% \usepackage{graphicx}

\subsection{Test set}
To build a test set, a very natural question is how many categories to select for the unseen set. We designed different sizes of unseen sets according to the number of images to be removed, and the details are shown in \Cref{tab:unseensplit}. 
%To this end, we  in table \ref{tab:unseensplit}, and 
An intuition is that higher unseen ratios lead to more stable and convincing evaluation results, but at the same time, higher ratios lead to a reduction of the training set resulting in lower model performance. The final results are shown in the  \Cref{tab:differentsplit}. In the case of too few unseen categories (0.1\%), the evaluation results fluctuate greatly due to randomness, but when the number of unseen categories exceeds 1\%, the evaluation results are gradually stable and can correctly reflect the goodness of the model, and in order not to further reduce the training data, we choose the case of 1\% unseen categories as our experimental setup.

\begin{table*}[ht]
\resizebox{\textwidth}{!}{%
\begin{tabular}{cllllllllllllll}
\toprule
\multirow{2}{*}{id} & \multicolumn{1}{c}{\multirow{2}{*}{$AR_{all}$}} & \multicolumn{1}{c}{\multirow{2}{*}{$AR_{kn}$}} & \multicolumn{2}{c}{0.1\%}                                     & \multicolumn{2}{c}{1\%}                                       & \multicolumn{2}{c}{5\%}                                       & \multicolumn{2}{c}{10\%}                                      & \multicolumn{2}{c}{20\%}                                      & \multicolumn{2}{c}{origin lvis val}                           \\ \cmidrule(l){4-15} 
                    & \multicolumn{1}{c}{}                            & \multicolumn{1}{c}{}                           & \multicolumn{1}{c}{$AR_{sn}$} & \multicolumn{1}{c}{$AR_{un}$} & \multicolumn{1}{c}{$AR_{sn}$} & \multicolumn{1}{c}{$AR_{un}$} & \multicolumn{1}{c}{$AR_{sn}$} & \multicolumn{1}{c}{$AR_{un}$} & \multicolumn{1}{c}{$AR_{sn}$} & \multicolumn{1}{c}{$AR_{un}$} & \multicolumn{1}{c}{$AR_{sn}$} & \multicolumn{1}{c}{$AR_{un}$} & \multicolumn{1}{c}{$AR_{sn}$} & \multicolumn{1}{c}{$AR_{un}$} \\ \midrule
1                   & 41.0                                            & 49.3                                           & 38.6                          & 53.9                          & 36.6                          & {\underline {47.5}}                    & 35.0                          & {\underline {45.0}}                    & 32.8                          & {\underline {44.2}}                    & 30.5                          & {\underline {42.7}}                    & 38.5                          & 47.4                          \\
2                   & 41.5                                            & 50.1                                           & 39.2                          & 53.9                          & 37.2                          & 47.9                          & 35.5                          & 45.5                          & 33.4                          & 44.6                          & 30.9                          & 43.2                          & 39.4                          & 48.0                          \\
3                   & 42.0                                            & 50.3                                           & 39.7                          & \textbf{54.3}                 & 37.7                          & \textbf{48.6}                 & 35.9                          & \textbf{46.2}                 & 33.8                          & \textbf{45.2}                 & 31.3                          & \textbf{43.8}                 & 39.8                          & \textbf{47.5}                 \\
4                   & 41.1                                            & 49.6                                           & 38.7                          & {\underline {53.4}}                    & 36.7                          & 47.6                          & 35.0                          & 45.1                          & 32.7                          & 44.3                          & 30.5                          & 42.8                          & 38.9                          & {\underline {47.0}}                    \\ \bottomrule
\end{tabular}%
}
\caption{\textbf{Comparison of results between different split ratios.}We train four identical models on the same training set with different random seeds, numbered 1, 2, 3, and 4. We evaluate them uniformly at the test set (lvis\_v1\_val + 10\%)  and calculate the AR according to the different divisions. For example, 0.1\% means that the 100 most uncommon classes are removed as unseen set $\mathcal{C}_{unseen}$, which accounts for about 0.1\% of all training sets. }
\label{tab:differentsplit}
\end{table*}

In \Cref{fig:gtandours}, we compare the prediction results of our method with the ground truth, and we can see that our method does detect many objects that are not in the ground truth. As we mentioned in Sec. 4.3 in the main text, if we use AP as a metric, these additional detected objects will be counted as false positives, resulting in extremely low AP, so AR is a more appropriate metric.
% \usepackage{booktabs}
% \usepackage{graphicx}

\section{Additional Ablations}
In the additional ablation experiments, we don't use example supervision, and the other settings are the same as in the main text.
\noindent \textbf{Training iterations.} Since our model is initialized with the already trained class-agnostic Mask2former before additional training, the experiment in  \Cref{tab:iters} is conducted in order to exclude the possibility that the performance gain comes only from the additional training iterations. Additional training based on the original mask2former does not effectively improve the model's open-world oriented segmentation ability, instead our method further improves performance with more training iterations.

\noindent \textbf{The maximum number of categories $C_{max}$.}
% Due to the computational cost is limited, the numbers of queries $N^{query}$ in prompt-based prediction branch is set to a fair number, $300$ for instance. In this case, the maximum number of categories $C_{max}$ that can be predicted per image becomes a critical hyperparameter, because a larger $C_{max}$ will reduced the number of the maximum prediction per category $K$ due to $N^{query} = C_{max} \times K$.
The maximum number of categories $C_{max}$ that can be predicted per image is a critical hyperparameter, because the number of negative class $C_{\text{neg}}$ is tightly relative to $C_{max}$, for instance, a large $C_{max}$ can bring a lot of negative class. Thus, we perform an experiment to investigate $C_{max}$, as shown in \Cref{tab:querynum}. Notably, the number of the maximum prediction per category $K$ is reduced when  $C_{max}$ increases, because the number of queries $N^{query}$ is fixed to $300$ and $N^{query} = C_{max} \times K$. The experiment results show that the model performance is improved steadily when $C_{max}$ is reduced. This trend reaches a state of saturation when $C_{max} = 15$. 

% \usepackage{booktabs}
% \usepackage{graphicx}
\begin{table}[]
\resizebox{\columnwidth}{!}{%
\begin{tabular}{@{}lcccccccc@{}}
\toprule
                            & Iters & $AR_{all}$    & $AR_{kn}$     & $AR_{sn}$     & $AR_{un}$     & $AR_{s}$      & $AR_{m}$      & $AR_{l}$      \\ \midrule
{%\color[HTML]{1F2329}
 M2F}  & 0     & 42.5          & 49.9          & 39.0          & 47.4          & 27.2          & 51.5          & 68.1          \\
M2F                         & 90k   & 42.1          & 49.8          & 38.5          & 47.1          & 27.0          & 50.9          & 67.5          \\
{%\color[HTML]{1F2329}
 Ours} & 90k   & \textbf{44.1}          & \textbf{51.9}          & \textbf{40.6 }         & \textbf{49.1}          & \textbf{28.9 }         & 53.3          & \textbf{70.6} \\
%{\color[HTML]{1F2329} Ours} & 360k  & \textbf{44.9} & \textbf{52.9} & \textbf{41.3} & \textbf{49.8} & \textbf{29.4} & \textbf{54.4} & 70.4         
\bottomrule
\end{tabular}%
}
\caption{\textbf{The results of different training iterations.} As the number of iterations increases the performance of our model can be further improved. The result in the first row is the Mask2former for 50 epochs of class-agnostic training, which is also the initialization weight of the class-agnostic baseline branch of our model. The second row shows that more training based only on the class-agnostic branch of Mask2former   \cite{cheng2022masked} does not improve the model and even degrades the performance. The last line reflects the effectiveness of our method.} 
\label{tab:iters}
\end{table}
\section{Implementation Details}

\begin{table}[]
\resizebox{\columnwidth}{!}{%
\begin{tabular}{c  c | c | c  c  c | c  c  c}
\toprule
$C_{max}$     &$K$             &  $AR_{all}$    & $AR_{kn}$         & $AR_{se}$     & $AR_{un}$         & $AR_{s}$      & $AR_{m}$          & $AR_{l}$   \\ \midrule
300           &1               &  43.0          & 51.2              & 39.4          & 48.1              & 27.3          & 52.6              & 69.6         \\
150           &2               &  43.8          & 50.9              & 40.5          & 48.5              & 28.0          & 53.3              & 70.9         \\
75            &4               &  44.0          & 51.2              & 40.6          & 48.8              & 27.9          & 53.8              & \textbf{71.2}\\
50            &6               & 44.2          & 51.6               & \textbf{40.8} & 49.1              & 28.5          & 53.9              & 70.5         \\
30            &10              & 44.1          & 51.9               & 40.6          & 49.1              & 28.9          & 53.3              & 70.6         \\
20            &15              &  \textbf{44.4}& 52.1               & \textbf{40.8} & 49.5              & \textbf{29.0} & \textbf{54.0}     & 70.6         \\
15            &20              &  \textbf{44.4}& \textbf{52.2}      & 40.6          & \textbf{50.0}     & \textbf{29.0} & 53.9              & 70.3         \\ \bottomrule
\end{tabular}%
}
\caption{\textbf{Varying the maximum number of categories $C_{max}$.} $C_{max}$ is the maximum number of categories that can be predicted per image and $K$ is the number of the maximum prediction per category. The number of queries $N^{query}$ in the prompt-based prediction branch equals $C_{max}$ multiply $K$. For a fair comparison, $N^{query}$ is fixed to 300 while the $C_{max}$ is varied.}
%\vspace{-1.5em}
\label{tab:querynum}
\end{table}

Our method can be trained from scratch or based on a pre-trained Mask2former. Except for the experiments on the full LVIS training set and COCO$\rightarrow$ LVIS, other experiments are based on the latter, and we find that the effect of freezing the backbone is better.

% \begin{figure*}[tp]
%     \centering
%     \includegraphics[width=\linewidth]{figs/results3.pdf}
%     \caption{\textbf{More qualitative results on few-shot segmentation.}The first two columns are the support image and its corresponding mask, and the last two columns are the segmentation results on the test images. We selected classes that are more difficult than those in \Cref{subsec:app} and  are all without annotations in the training time.}
%     \label{fig:fewshot2}
% \end{figure*}

\noindent \textbf{Prompt learning mechanism.} For prompt extraction and learning, we filter out the mask annotations with an area smaller than 100 on $1024 \times 1024$ images. This is caused by the mechanism of mask-attention. The area of the small mask on the smaller feature map will become 0, forcing the mask-attention to focus on all regions of the whole picture, which cannot effectively extract the corresponding object information, resulting in the quality of the prompt cache being poor.

\noindent \textbf{Open-vocabulary/Few-shot Segmentation.} Unlike the above, where the prompt extraction branch and the prompt prediction branch are used as auxiliary training modules, here we need to use the prompt prediction branch directly for inference, and the class-specific prompt will use the reference attention to combine the information from the class-agnostic branch to segment the objects in the image corresponding to the class. In addition, there are some differences in the training process, we introduce the image-level label of the seen $\mathcal{C}_{seen}$ into the prompt prediction branch, and the candidate query of this part only receives the supervision of the classification loss, but not the supervision of the mask. In order to make each category query more accurate in locating objects in the corresponding category and to facilitate the constraint by classification, we set $K$ to 1, i.e., each category has only one query for prediction. For open-vocabulary segmentation, we replace all learnable class-specific embedding $\mathbf s$ with fixed CLIP   \cite{radford2021learning} text embedding and drop the original prompt extraction branch, and the model can do open-vocabulary segmentation after training. For the few-shot segmentation, we keep the prompt extraction branch. A small number of support images are used as input, and the model also performs a momentum update on the prompt cache and then uses the prompt to segment the test images.%\Cref{fig:fewshot2} shows more visualization of the results.